# A self-portrait of young Leonardo

**Amelia Carolina Sparavigna**
**Dipartimento di Fisica**
**Politecnico di Torino**

One of the most famous drawings by Leonardo da Vinci is a self-portrait in red chalk, where he looks quite old. In fact, there is a sketch in one of his notebooks, partially covered by written notes, that can be a self-portrait of the artist when he was young. The use of image processing, to remove the handwritten text and improve the image, allows a comparison of the two portraits.

A special event at the Reggia di Venaria, Torino, is the exhibition "Leonardo, the Genius the Myth", from November 2011 to January 2012 [1]. The famous self-portrait in red chalk (Fig.1) will be on display in this exhibition along with works by other artists inspired over the centuries by the Leonardo's genius. The exhibition, as a journey through Leonardo da Vinci's works, includes several original drawings. The exhibition room was designed by Dante Ferretti, two times awarded by the Academy for Best Art Direction, who created a special shrine to contain the Leonardo's drawings. Besides the famous self-portrait, the visitors can see the Codex on the Flight of Birds, opened at the pages where, according to Carlo Pedretti, an Italian historian expert on the life and works of Leonardo, there is a self-portrait of the young genius [2,3]. As the self-portrait in red chalk, the Codex on the Flight of Birds is held at the Biblioteca Reale of Turin. The code is relatively small, dated approximately 1505.

As we can see for Figure 2, the drawing in the Codex had been partially hidden by Leonardo, who wrote on it. Therefore, this portrait remained unappreciated for 500 years before being noted by Piero Angela, an Italian scientific journalist. On Saturday February 27, 2009, during a prime-time entertainment show of the RAI broadcaster on history and science [4], Piero Angela explained how he noted this portrait of a young man. Angela said that when he was observing a copy of the Codex on the Flight, he noticed what looked like a nose underneath the text (Fig.2). The fact that the drawing was made by a left-hand artist, as the directions of the sketching lines indicate, reinforced the supposition of a self-portrait of the artist. Moreover, comparing the drawing with a Leonardo self portrait of c. 1512-15, Angela told that the two men were looking like brothers.

As previously told, a well-known researcher on Leonardo studies, Carlo Pedretti, agrees in considering the image as a self-portrait. Carlo Pedretti was the first to propose a "restoration" of this drawing, by removing the handwritten words on a photographic plate. In 2009, Piero Angela presented the digital restoration of the portrait, obtained by enhancing the red-chalk sketch on a high resolution digital image. The graphic designers gradually cancelled the text revealing the drawing beneath: after months of micro-pixel work, the portrait of this Renaissance man appeared (see the result at the site [5]). As discussed in [6,7], this was a very important discovery, demonstrating that the image processing is a fundamental tool for a new kind of restoration, not made on the document itself but on its digital image. The digital restoration can give excellent results, important for studies of art and palaeography.

In Ref.6, I have proposed a simple approach based on interpolation with nearest neighbouring pixels, for the pleasure to repeat the discovery of a Leonardo self-portrait. Let me shortly repeat what is the procedure of such digital restoration (for more details [6]). Each pixel of a colour digital image can have red, green and blue tones (RGB) with numerical values ranging from 0 to 255. Because the portrait is in red-chalk and the writing almost black, we can choose a threshold value to remove the darkest pixels and replace them with white pixels. The new image is shown in Fig.3, up-right panel, and we can further work on it. In this image, the hand-written text is white. Instead of working on the white pixels manually, that is, changing them pixel by pixel, we can prepare an algorithm as follows. Let us consider a white pixel: we replace it if three pixels in its nearest neighbour are not white. The new pixel has the colour tones given by the averaged values of these three pixels. This reconstruction procedure is iterated until almost all the white pixels had been removed and we have the portrait as in Fig.3, down-left panel [6].

Here I am proposing a further processing of this image with a wavelet filtering program [8]. Such filtering allows the adjustment of the image features at several different scales. The result after processing with the freely available software Iris [9], is displayed in the down-right panel of Figure 3 and in Figure 4, to have better resolution of the portrait. Besides Iris, another freely available image processing tool is GIMP [10]. It is a very interesting software that allows the merging of images on several layers, each having its proper transparency level. Then GIMP gives us another quite interesting possibility, that of working on the two portraits, that of Fig.2 and Fig.4, because we can superimpose them. The final result is shown in Figure.5: the two faces seems quite coincident: in particular, the relative distances of eyes, nose and mouth are the same. Moreover, this processing makes Leonardo look younger.

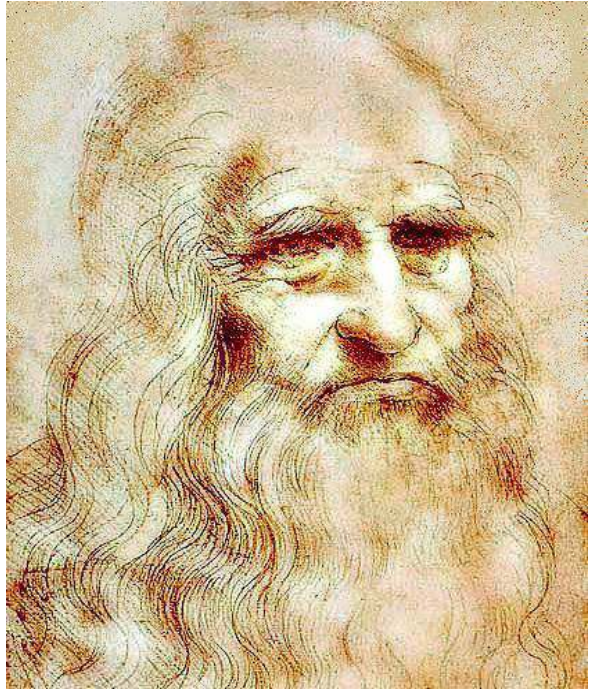

Fig.1 Leonardo da Vinci self-portrait in red chalk, held at the Biblioteca Reale of Turin

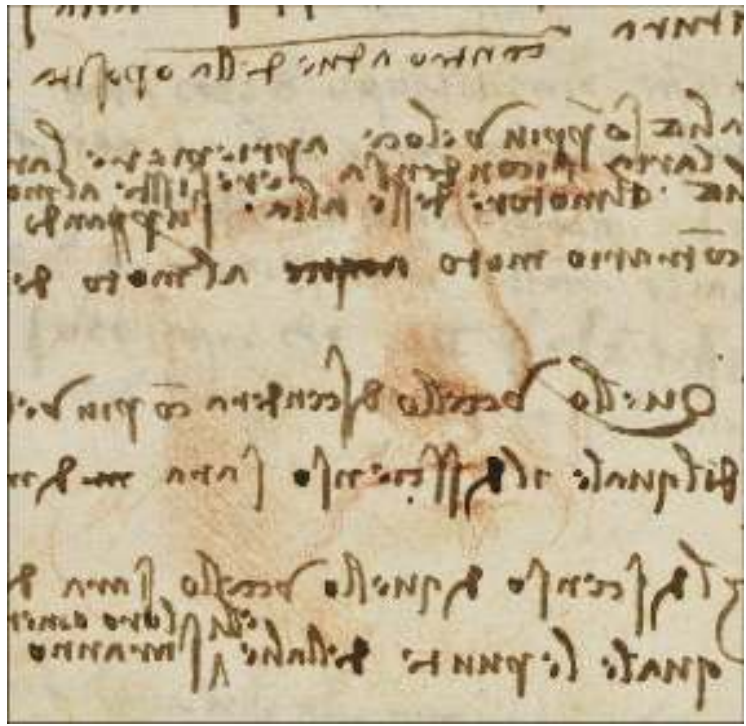

Fig.2. Piero Angela initially discovered what looked like a nose in a page of the Codex on the Flight of Birds. If we remove the writing, the portrait appears.

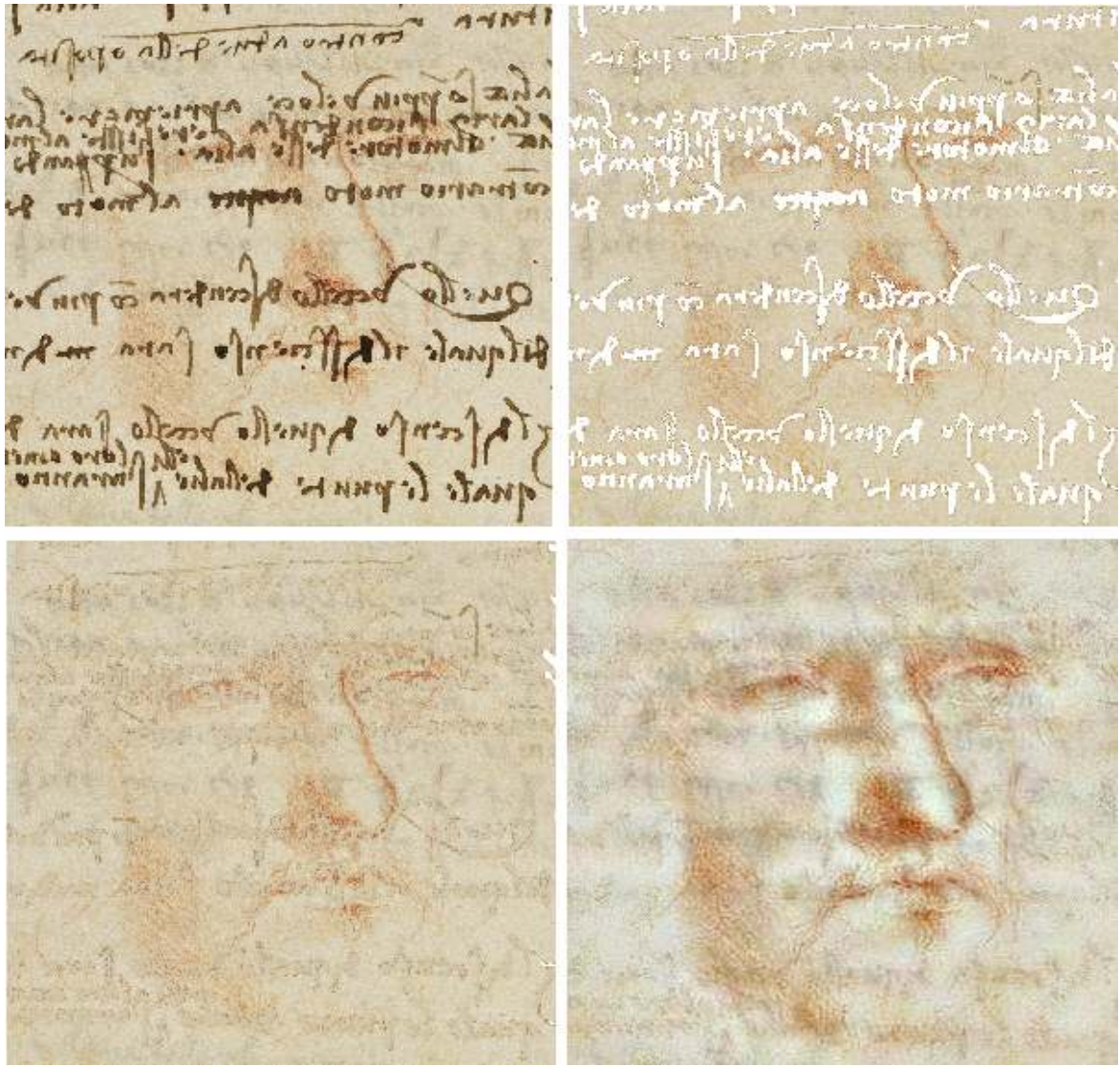

Fig.3. The drawing has red tones, the handwritten text is black (original, up-left). We can replace the black pixels with white pixels (up-right panel). Then we can work on the white pixels, replacing them with proper colour tones given by the neighbouring pixels (down-left panel). A further processing of this image with a wavelet filter gives the image in the down-right panel.

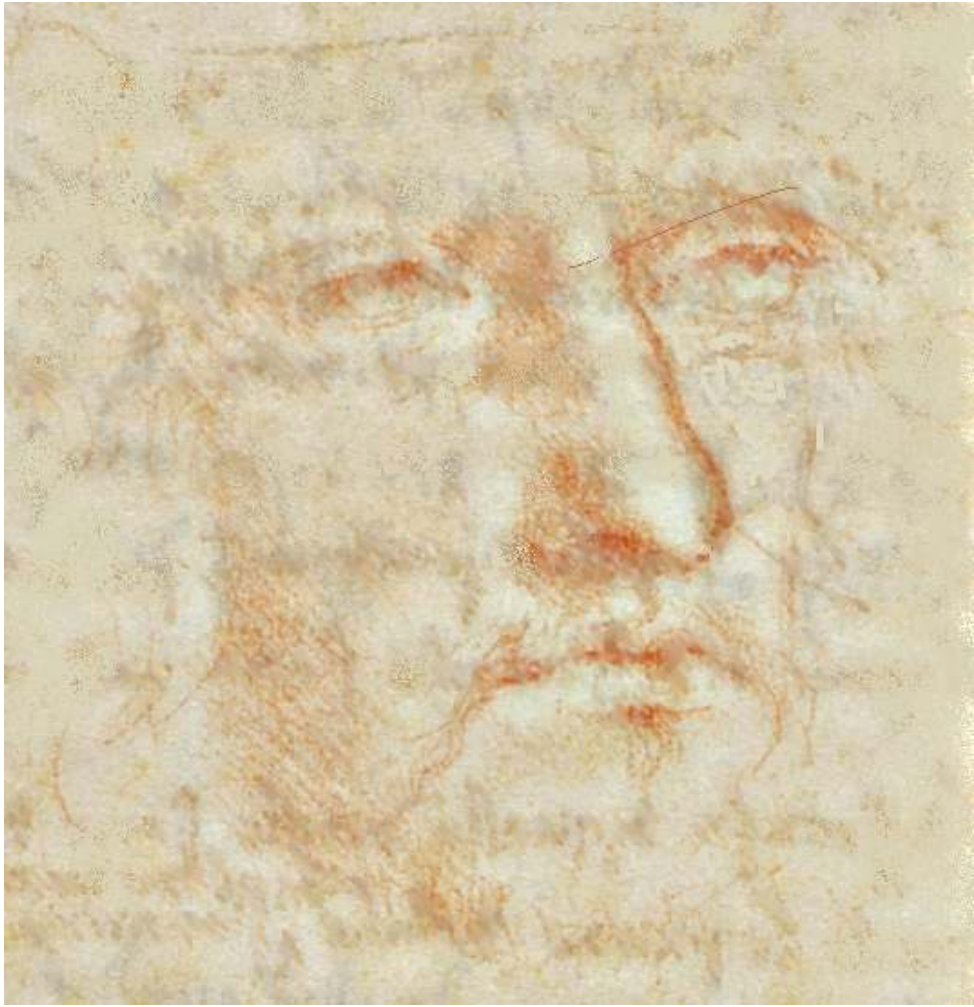

Fig.4 A high resolution image of the down-right panel of Figure 3.

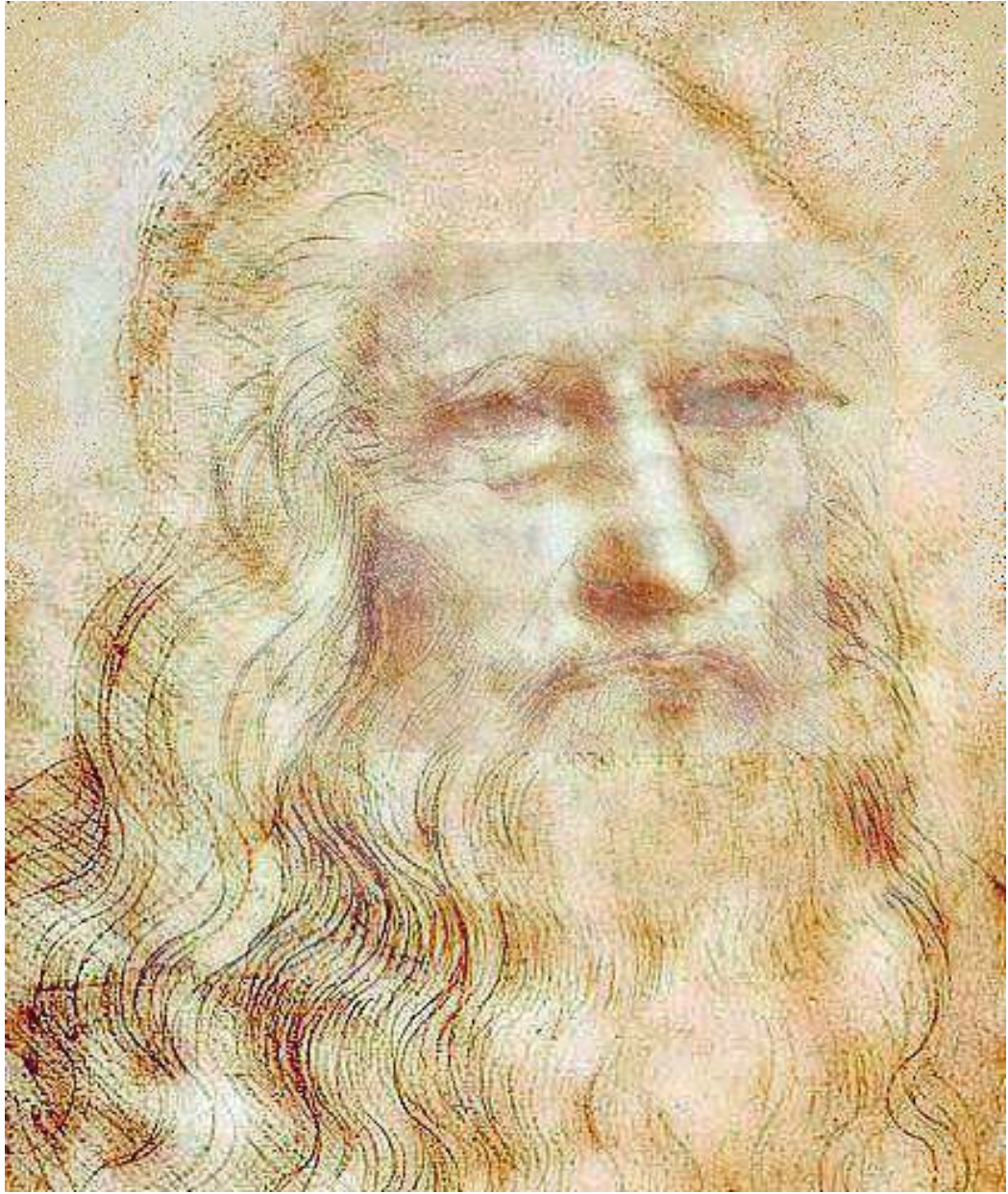

Fig.5 GIMP allows the merging of the two portraits, that of Figure 1 and Figure 4. The final result shows that the two faces seems quite coincident in a younger Leonardo da Vinci.